\pdfoutput=1

\documentclass[11pt]{article}

\usepackage{naacl2021}

\usepackage{times}
\usepackage{latexsym}
\usepackage{amsmath}
\usepackage{graphicx}

\usepackage{subfigure}

\usepackage[T1]{fontenc}

\usepackage[utf8]{inputenc}

\usepackage{verbatim}
\usepackage{amssymb}

\usepackage{microtype}

\title{LayoutXLM vs. GNN: An Empirical Evaluation of Relation Extraction for Documents}

\author{Hervé Déjean, Stéphane Clinchant, Jean-Luc Meunier \\
  Naver Labs Europe \\
  \url{www.europe.naverlabs.com}\\
  \texttt{firstname.lastname@naverlabs.com} 
}
\begin{document}
\maketitle
\begin{abstract}
This paper investigates the Relation Extraction task in documents by benchmarking two different neural network models: a multi-modal language model (LayoutXLM) and a Graph Neural Network: Edge Convolution Network (ECN). For this benchmark, we use the XFUND dataset, released along with LayoutXLM. 
While both models reach similar results, they both exhibit very different characteristics.
This raises the question on how to integrate various modalities in a neural network: by merging all modalities thanks to additional pretraining (LayoutXLM), or in a cascaded way (ECN).
We conclude by discussing some methodological issues that must be considered for new datasets and task definition in the domain of Information Extraction with complex documents.
\end{abstract}

\section{Introduction}

The last years have seen numerous publications in the Natural Language Processing (NLP) community addressing the problem of Information Extraction from complex documents. In this context the term "complex documents" means that considering a document as a mere sequence of sentences is too simplistic: the layout (position of the elements in the page) carries meaningful information. Typical examples are forms and tabular content.
Historically, such documents have been addressed by the Computer Vision (CV) community, and formalized as the Document Layout Analysis task, defined as the process of {\it identifying} and {\it categorising} the regions of interest in a document page.

Recent publications have addressed this task by adapting pre-trained language models enriched with geometrical information (text position in the page). 
The first attempt went toward multi-modal models combining image (scanned page) and text (provided by OCR) such as \citet{Katti2018ChargridTU}. A major step was the adaption of  language models with 2D positional information (text position in the page) with \citet{Xu2020LayoutLMPO,Xu2021LayoutLMv2MP,  Garncarek2021LAMBERTLL, Hong2021BROSAP}. This approach requires an additional pre-training step for integrating this new 2D positional information (millions of pages are usually used). Building on this method, various approaches have been released combining or not image information. 


Beyond the traditional word/region categorisation task (seen as named entity recognition by layoutLM like models), the Relation Extraction (RE) task is also tackled with these multi-modal architectures. This task aims at linking two related textual elements such  as the question field and the answer field in a form.
Evaluation shows that traditional language models à la BERT perform very badly, and that the addition of 2D positional embedding is of key importance.  

In this paper, we investigate whether the use of layoutLM like models is relevant for this RE task and benchmark against a Graph Neural Network model: Since, in some recent work such as  \citet{Hong2021BROSAP,Zhang2021EntityRE}, the decoder part, based on fully connected graph, allows for good performance, we assume that GNN naturally is well adapted for this RE task.
We selected the Edge Convolution Network (ECN) proposed by \citet{clinchant2018comparing} and benchmark it using the XFUND dataset \cite{Xu2021LayoutXLMMP}. Initially designed for layout segmentation, we adapted it for the Relation Extraction task.



\section{Models}
We present the two models we are using in this benchmark: LayoutXLM  and ECN.

\subsection{LayoutXLM}
LayoutXLM \cite{Xu2021LayoutXLMMP} is a multilingual version of LayoutLMv2 \cite{Xu2021LayoutLMv2MP} using  XLM-RoBERTa language model \cite{Conneau2020UnsupervisedCR} instead of UniLMv2. 

LayoutXLM uses four types of  embeddings: the usual text embedding and 1D positional embedding, an additional 2D embedding corresponding to the top left and bottom right coordinates of a token bounding box plus the height and width of the bounding box. Eventually, an image embedding of the page regions using a Resnet backbone is also used. They are added the same way as the 1D positional embedding. The new model is trained with  11 million pages from the IIT-CDIP dataset using adapted pre-training tasks. 

For the Relation Extraction task, LayoutXLM builds all pair of possible entities. They concatenate the  \textit{first token vector in each entity and the entity type embedding obtained with a specific type embedding layer. After respectively projected by two FFN layers, the representations of head and tail are concatenated and then fed into a bi-affine classifier.} \cite{Xu2021LayoutXLMMP}.
The entity type is provided by the ground truth, and considerably facilitates the RE task as we will see Section~\ref{Evaluation}.
The number of parameters of the LayoutXLM-base model is 354M, the specific decoder accounts for  3M parameters.
In order to add the additional task settings (see Section \ref{sec:xfund}), we simply modify this decoder part (by deleting the label representation or adding an embedding dimension when considering 'Other' entities).



\subsection{Edge Convolution Network (ECN)}
\citet{clinchant2018comparing} proposed a graph neural network architecture which departs from traditional GNN: 
they were the first showing that it is important for these datasets (documents) to distinguish between the node representation and the neighbourhood representation (approximated with a residual connection in \citet{bresson2017residual}, and latter named \textit{Ego- and Neighbor-embedding Separation} in \citet{H2GCN}).

This method uses  some prior in order to build the  graph and to create edge features.
The graph is built using the page elements as nodes and the edges are created using the line-of-sight strategy (two nodes are neighboured if they see each other). 
The node embedding corresponds to the concatenation of its textual embedding (we used XLM-RoBERTa and Bert-base-multilingual) and its geometrical embedding $E=(t,x_{0},y_{0},x_{1},x_{1},w,h)$. The textual embedding $t$ is generated using a language model (monolingual or multilingual, see Section~\ref{exp}), and corresponds to a pooling (mean) of the last layer embedding\footnote{The CLS token provides worse result.}. The geometrical embedding  is built using the 6 usual values: top left, bottom right coordinates, the  width and height of the bounding box containing the text (values are normalised as in \citet{Xu2021LayoutXLMMP}).
Similarly to \citet{Zhang2021EntityRE}, edges features are generated by using geometrical properties of the two nodes of an edge. We modified the original ECN as follows:

\textbf{Fully Connected Layer:} While this line-of-sight graph generates a good  prior  for the initial clustering task, it is not appropriate for the Relation Extraction task where some linked elements are not necessary neighbours. We simply use the same approach  as in \citet{Hong2021BROSAP, Zhang2021EntityRE} by adding a last layer representing a fully connected graph. The edges of this graph are represented by concatenating the embedding of both edge nodes provided by the previous layer.

\textbf{Edge embedding:} While edge features are used, their representation is not learnt: in our version, an edge representation is learnt for each layer $n$ by projecting the representation from layer $n-1$ using a Feed-Forward Network.

\begin{table*}[ht]
\centering
\small
\begin{tabular}{|l||c|c|c|c|c|c|c|c||c||c||c|}
\hline
\textbf{Model}       &     \textbf{Entities}  & \textbf{ZH} & \textbf{JA} &\textbf{ES} & \textbf{FR} &\textbf{IT}  & \textbf{DE} &  \textbf{PT} & \textbf{AVG1} &\textbf{EN}& \textbf{AVG2}\\
\hline
\hline
\multicolumn{12}{|c|}{\textbf{With label (official setting)}}\\
\hline
\hline
LayoutXLM-base   & HQA   & 70.73  &69.63  &  69.96      & 63.53  &64.15 & 65.51 & 57.18  & 65.81 &54.83 &  64.44  \\
LayoutXLM-large   & HQA   & 78.88     & 72.55 &    76.66  &   71.02     &76.91 & 68.43& 67.96 &73.20    &64.04  &72.06 \\
ECN  no text       & HQA  &  87.4    & 80.16     &   78.60   & 84.80     &80.0   &77.33&71.30  & 79.94  &83.95  &\textbf{80.44} \\

\hline
\hline
\multicolumn{12}{|c|}{\textbf{Without label}}\\
\hline
\hline
LayoutXLM-base      & HQA   &      68.10 & 65.69  & 67.71 &59.35   & 65.69 &61.39 & 54.22    &  63.13   &47.38  & 60.64\\
ECN  bert-multi    & HQA  & 79.12     &  69.75    & 70.06    & 77.18 &73.80   & 59.72& 60.18& 69.75 & 73.20  & \textbf{70.18} \\
\hline
LayoutXLM-base  & OHQA  & 69.53     & 57.21 & 62.78   & 55.00  & 65.00 & 57.89& 69.73 & 62.45  & 55.00   & 61.52\\
ECN  XLM  & OHQA  &  72.13    & 59.62     & 57.22  & 69.58 &66.45   & 56.80& 51.68 & 61.91  &68.00  &62.67  \\
ECN bert-multi  & OHQA  & 73.52     &58.42      &     59.86        &  69.82      & 66.78  & 60.90&54.60  & 63.41  & 69.22 & \textbf{64.14} \\

\hline
\end{tabular}
\caption{\label{mono}XFUND Relation Extraction task. Monolingual. Average of 5 runs except for LayoutXLM, official setting.}
\end{table*}

\begin{table*}[ht]
\centering
\small
\begin{tabular}{|l||c|c|c|c|c|c|c|c||c||c||c|}
\hline
\multicolumn{12}{|c|}{\textbf{ \textsc{Multilingual}}}\\
\hline
\textbf{Model}       &     \textbf{Entities}  & \textbf{ZH} & \textbf{JA} &\textbf{ES} & \textbf{FR} &\textbf{IT}  & \textbf{DE} &  \textbf{PT} & \textbf{AVG1} &\textbf{EN}& \textbf{AVG2}\\
\hline
\hline
\multicolumn{12}{|c|}{\textbf{With label (official setting)}}\\
\hline
\hline
LayoutXLM-base  & HQA    & 82.41 &81.42  &81.04 &    82.21    &83.10   &78.54 &70.44  &   79.88   & 66.71  & 78.23    \\
LayoutXLM-large & HQA    &  90.00 &86.21  &85.92   & 86.69 &  86.75   &82.63   & 81.60  &  85.79  & 76.83     &84.58 \\
ECN no text      & HQA    & 90.82     &    86.67  &  89.66    & 92.22  & 86.08 &85.72&81.64  & 87.55  &89.27  & \textbf{87.76}  \\
\hline
\hline
\multicolumn{12}{|c|}{\textbf{Without label}}\\
\hline
\hline
LayoutXLM-base  & HQA &  83.28 & 79.30 & 81.63   & 80.46 & 79.58  &75.61 & 70.03 &78.56  &65.77 &76.96      \\
ECN-XLM & HQA  &  82.88 & 75.95 & 79.14  &83.96 &77.60&74.04  & 69.43  &77.57 &81.09  & \textbf{78.01}   \\
\hline
LayoutXLM-base   & OHQA &  77.97 & 69.45 &75.11  &75.60 & 74.91  &70.79 & 64.52 &72.62  &63.00 &71.42      \\
ECN-XLM& OHQA  & 77.88 & 64.90 & 69.61  &77.00 &73.60&69.11   & 63.53       &70.80 & 77.41 &71.63   \\
ECN-bert-multi    & OHQA  & 79.28& 66.62 & 72.62  &77.02&73.08&71.24  &64.51    &72.05&82.06 &\textbf{73.30}  \\
\hline
\end{tabular}
\caption{\label{multi}XFUND Relation Extraction task. F1 score. Using all languages for training. Average of 5 runs except for LayoutXLM, official setting.}
\end{table*}

\section{The XFUND Dataset}
\label{sec:xfund}
The XFUND dataset \cite{Xu2021LayoutXLMMP} is a multilingual extension of the English FUNSD  \cite{funsd}. It contains 8 sub collections in 8 languages (the English one corresponding to FUNSD). 
In this dataset, results for each language is provided as well as a multilingual setting: all the 8 languages are used for training and evaluation is performed for each language individually.
A page is represented as a set of textual entities categorized in 4 classes: header (H), question (Q), answer (A) and other (0). 
XFUND proposes two tasks called Semantic Entity Recognition (SER) and Relation Extraction (RE). 
The SER task consists in tagging the words in the 4 entity classes (formalized as a Named-Entity-Recognition task).
The RE consists in linking two related entities, and is limited to the question/answer relation, ignoring the header relation.
We now discuss some issues with the XFUND dataset, its RE task and its impact on the evaluation.

\textbf{512 curse:} As for many language models, LayoutXLM can only process sequences of limited length (512). Hence documents longer than this length are split. This has a strong impact on the initial relations since LayoutXLM simply ignores relations between two 'sub-documents'. Table~\ref{512} in appendix shows the impact of this split: 12\% of the relations are lost, and we evaluated the impact on the F1 score higher than 3 points (impacting the recall) as shown by the last row. We initially did not think it would have such an impact on the LayoutXLM model and this strongly  biases the comparison in favour of LayoutXLM. How to solve this methodological problem must be clearly decided in the future.

\textbf{Artificial Setting:} The RE task is performed on the Header/Question/Answer (hereafter HQA) entities only ignoring the O(ther) entity. Furthermore the label of each entity is also known.  We use this setting as well but add more realistic ones. We first remove the label information (with/without label row), and secondly the Other entities are added (OHQA setting, Entities column). This last setting is still not totally realistic since the ground-truth is used for identifying the entities. A completely realistic setting is beyond the scope of this short paper. 

\textbf{Validation Set:} Neither XFUND nor FUNSD provide a validation set. Previous work usually stops after $N$ iterations (typically 100 for FUNSD). The training settings for LayoutXLM are not provided. Since a parameter finetuning is required, we consider the test set as \textit{de facto} validation set. Used for both methods (LayoutXLM and ECN),it does not hurt the benchmarking. Alternatively we could have created a validation set from the training set, but none of the previous work has done this. This leads to a high standard deviation for both systems (see Table~\ref{512} in appendix). 

\textbf{Adding the English FUNSD dataset:} The English version (FUNSD) is not included in XFUND and we converted it into the proper format. Some results are a bit weird for this language: ECN performing far better than LayoutXLM. It might be due to an issue in our own conversion, but also because the guidelines for both datasets do not follow the same rule. So we provide two general results: without (column AVG1) and with (AVG2) English.

\subsection{Experimental Setup}
\label{exp}
For LayoutXLM, we finetuned the learning rate (1e-5 to 5e-5) and the batch size (1 to 6 documents), and the number of epochs (10 to 200 depending of the setting). Only the base model has been released, we are thus not able to provide evaluation for LayoutXLM-large.
For ECN, we finetuned the dimension for the node and edge representation, (128, 256), the number of layers (4-8), the number of stacked convolutions (4-5), and the number of epochs. The batch size was fixed to 1 document. 
Appendix \ref{sec:appendixE} describes the final setting for both models and the parameters for each architecture.
Experiments were conducted using a Tesla V100 with 32G memory. Training is 30-40\% faster with ECN (monoling. 30mins/50mins; multiling. 4hs/6hs).


\section{Evaluation}
\label{Evaluation}
The results  shown Tables~\ref{mono} and \ref{multi} correspond to the mean of 5 runs randomly initialised. The official results from \citet{Xu2021LayoutXLMMP} use the seed 43. 

\textbf{Monolingual setting:} Table~\ref{mono} shows the results for the evaluation with the monolingual setting. We clearly see that the use of the entity label as done in \citet{Xu2020LayoutLMPO} is an artificial setting since the task is better performed without textual information: The ECN (no text) performs 14 points higher just using the geometrical embeddings. Without label, and ignoring the Other entities, ECN still performs  better. In the final setting (no label and all 4 entities), both models are very close. By selecting the BERT-base-multilingual language model instead of XLM-RoBERTa-base for generating the textual embeddings, ECN becomes slightly better.   

\textbf{Multilingual setting:} Table~\ref{multi} shows the results for the multilingual setting (all languages used for training). As in \citet{Xu2021LayoutXLMMP}, it shows that this setting is very efficient: the transfer among languages works. Again, the use of labels makes the textual embeddings useless. When the label is not used, both models perform equally (no statistical difference using a paired or unpaired t-test with p < 0.05).
For both settings, we can note that some languages are better covered by one model (PT and LayoutXLM or FR and ECN). It is also not clear why the BERT-base-multilingual performs for almost all languages better than XLM-RoBERTa.  

\section{Discussion}
\label{sec:discussion}
The results show that in the most generic setting (no label, all entities) both models perform similarly. The ECN model has some advantages: first it is 70 times smaller than LayoutXLM, it does not require images as input, nor specific "middle"-training to integrate non textual information, and can be seen as a specific decoder.
On the other hand, LayoutXLM seems to better react when the size of the dataset increases: in the most generic setting (no label, all 4 entities) it catches up  the ECN performance. Furthermore, we were not able to test the large version of LayoutXLM (not released), which really improves results in the official RE task.
This study raises an interesting point when dealing with multi-modality: both systems integrate textual and geometrical information is a very different manner. In LayoutLM/XLM, geometrical/image embedding are integrated by pre-training an initial language model, while with ECN, the language model is simply used for generating the textual embedding and ECN is more focused on the geometrical aspect of the problem. One advantage of this configuration, not shown in this paper, is that in the monolingual setting, a specific language model may improve the results further. For instance with a Chinese Bert model (chinese-bert-wwm), you gain 3 points F1.  We also tried to combine a language model as trainable backbone and ECN as decoder, but the results were disappointing (despite \citet{Zhang2021EntityRE} results), and we are still investigating this point.

We have found many methodological issues in using the XFUND dataset: no validation set, missing relations due to the 512 max length split. One additional issue is the fact that the reading order is provided by the ground-truth is perfect, while \citet{Hong2021BROSAP} has shown that the token order matters a lot for language models. We hope that the future datasets will fix these points. 



\section{Conclusion}
In this paper, we compared a multi-modal language model (LayoutXLM) and a graph convolution model (ECN) for Relation Extraction in documents using XFUND. We compare them by using the official RE task and introduce more realistic settings. In the official setting (with label) we show that the textual information is useless to perform the task. For the more realistic settings, both models perform similarly. The ECN model has the advantage of not requiring any pretraining. but requires some minor (document generic) prior knowledge.  
These results question the way the multi-modality is achieved with the pre-trained language models: these models are able to leverage this multi-modalities, but  do not seem to use them yet in an optimal  manner. We think this multi-modality problem still needs investigation, and not only in the mainstream direction.
Finally we raise methodological issues with XFUND, hoping that the next dataset releases will take care of them.



\bibliography{doc,mibib}
\bibliographystyle{acl_natbib}

\appendix
\section{Appendix A: Illustration of the XFUND dataset}
\begin{figure*}
    \centering
    \includegraphics[width=160mm]{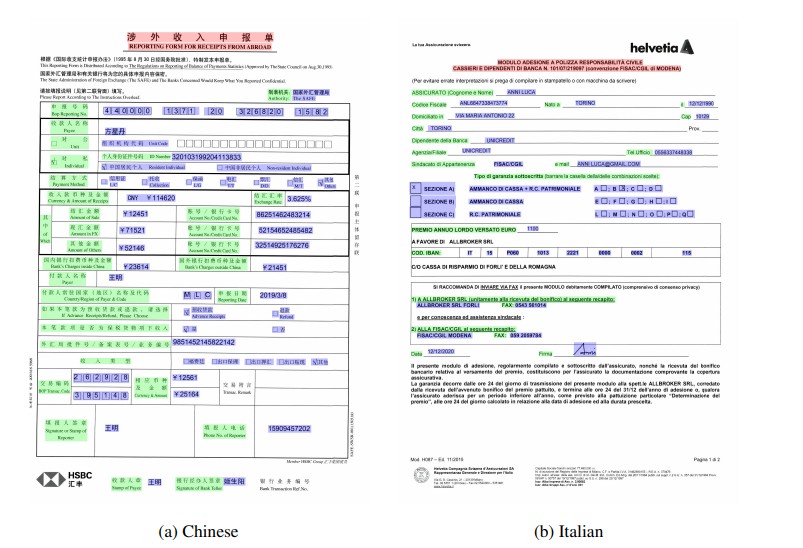}
    \caption{2 examples of the XFUND dataset \cite{Xu2021LayoutXLMMP}.}
    \label{fig:my_label}
\end{figure*}

Figure~\ref{fig:my_label} shows two examples of the XFUND dataset,  where red denotes the headers, green denotes the questions  and blue denotes the answers. The Relation Extraction consists at linking answers to the right question.

\section{Appendix B: Digging into ECN }
We present here the updated equation of our modified version of ECN (edge embeddings):
\begin{flalign}
\label{equationh0}
& h_{i}^{(0)} = x_{i}  \\
& \phi_{ij}^{(0)} = \phi_{ij}  \\
\label{eqn}
& \mu_{i}^{(l+1)} = \Phi^{(l)}h_{j}^{l} \\
\label{equationhi}
& \gamma^{l+1}  =\operatorname{\oplus}_{k=1}^{K} \sum_{j \in \mathcal{N}(i)} W^{k} (\Psi^{l}_{ij} e_{ij}) * \mu_{j}^{l+1}   \\
\label{equationN}
& h_{i}^{(l+1)} = \sigma (\mu_{i}^{l+1}  \oplus \gamma^{(l+1)} )\\
\label{equationFinal}
\end{flalign}
where $W$  is the weight  matrix for edge features, $\Phi$ and $\Psi$ feedforward network used for updating the node and edge representation at layer $l+1$,  $ \oplus$ the  concatenation operation,$K$ the number of stacked and $\mathcal{N}$ the neighbourhood of node $\it{i}$.
In Equation~\ref{equationhi} the node representation for the layer $l+1$, in Equation~\ref{equationN}, the neighbourhood representation in computed by  concatenating $K$ convolutions, and finally Equation~\ref{equationFinal} concatenates the node representation and its neighbourhood representation to produce the final representation. This set of equations allows for keeping separated the node and neighbourhood representations. 

\section{Appendix C: ECN Graph}
\begin{figure}
    \centering
    \includegraphics[height=80mm]{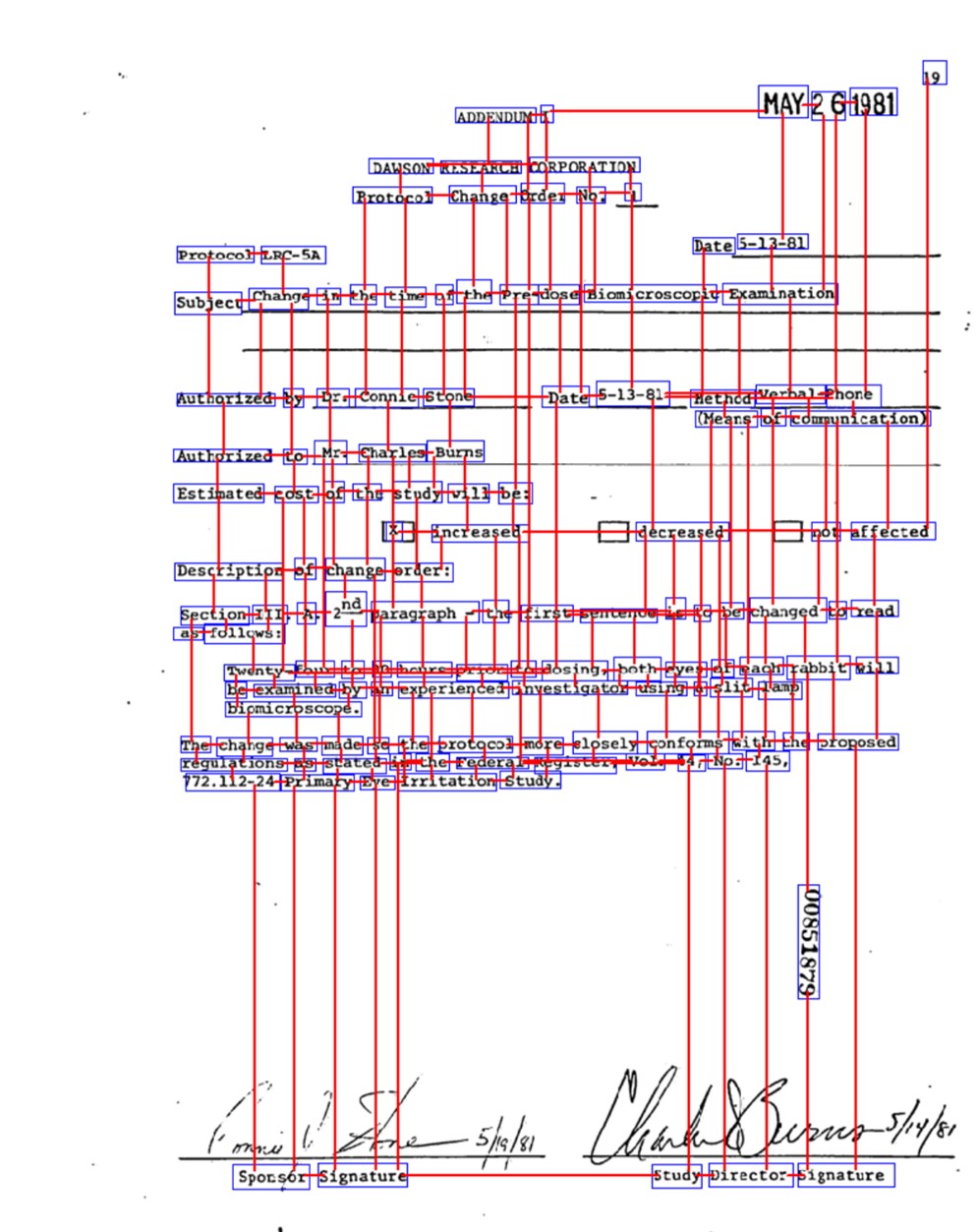}
    \caption{A page as a graph: nodes (blue) represent textual page objects, and edges (red) represent spatial relations between nodes using the line-of-sight strategy.}
    \label{fig:pageasagraph}
\end{figure}

Figure~\ref{fig:pageasagraph} shows an example of an initial graph used by the ECN model.
\label{sec:appendixA}

\section{Appendix D: Edge embeddings}
\label{sec:appendixD}
\begin{figure*}
    \centering
    \includegraphics[width=60mm]{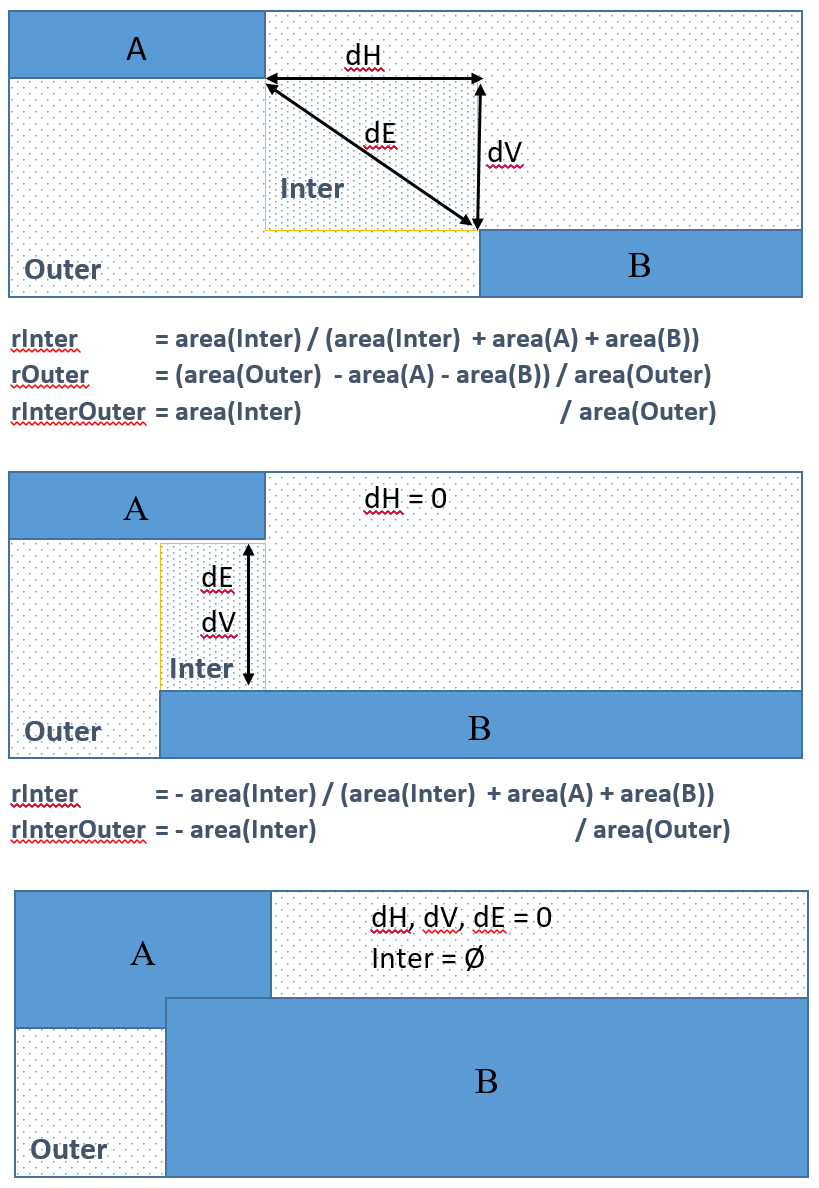}
    \caption{Illustration of the geometrical edge features used by the ECN model.}
    \label{fig:edgefeat}
\end{figure*}

We use 14 features to define  an edge:
\begin{itemize}
 \setlength\itemsep{0em}
\item 3 distances: horizontal, vertical, Euclidean between the closest point of the bounding boxes of the two blocks, or 0 if overlap as explained in the schema below.
\item 3 area ratios: rInter, rOuter, rInterOuter, as explained in schema below
\item x1, x2, y1, y2 of source node
\item x1, x2, y1, y2 of target node
\end{itemize}
We distinguish several situations: no overlap, overlap of projections on 1 axis, true overlap, as shown Figure~\ref{fig:edgefeat}.

\section{Appendix E: Experimental Settings}
Table~\ref{settings} provides the list of hyperparameters we finetuned for each model as well as the selected values. As mentioned, no validation set is provided, so the finetuning has been done with the test set for both models and both settings (monolingual and multilingual).

\label{sec:appendixE}
\begin{table*}[ht]
\centering
\small
\begin{tabular}{|l|c|c|c|}
\hline
\textbf{Model} & Parameter grid  & Selected (monolingual)  & Selected (Multilingual)\\
\hline
\multicolumn{4}{|l|}{\textbf{\citet{Xu2021LayoutXLMMP}  }}\\
\hline
Batch size (document level) &1-8  & 6 & 2 \\
Epochs  & 10-50-100-150-200 & 150 & 50 \\
Learning rate & $1e^{-5}-5e^{-5} $ & $3e^{-5}$  & $1e^{-5}$ \\
\hline
Number of Parameters (million)  & & 354 &354\\
\hline 
\multicolumn{4}{|l|}{\textbf{\citet{clinchant2018comparing} modified }}\\
\hline
Batch size  &1 & 1& 1\\
Epochs  &100-200-400 &400 &400\\
Learning rate &$5e^{-3}$-$5e^{-4}$ &$5e^{-4}$ &$5e^{-4}$\\
Node dimension &128-256 &128 &256\\
Edge dimension &64-128 & 128&128\\
Layers &4-6-8 & 6& 6\\
Stacked convolutions &4-6-8  &6 &8\\
\hline
Number of Parameters (million)  & & 1.2 & 5.5\\
\hline
\end{tabular}

\caption{\label{settings}Parameter grid used for \cite{Xu2021LayoutXLMMP} and \cite{clinchant2018comparing} }
\end{table*}

\begin{table*}[ht]
\centering
\small
\begin{tabular}{|l|c|c|c|c|c|c|c|c|c|c|}
\hline
\textbf{Language}                  & \textbf{EN} &\textbf{ZH} &\textbf{JA} & \textbf{ES} & \textbf{FR} & \textbf{IT} & \textbf{DE} & \textbf{PT} & \textbf{Total}\\
\hline
\multicolumn{10}{|l|}{Training set} \\
\hline
\# relations (full documents)     &  3129 &4621  & 3819&4239 &3425 &4927 &3982 & 5414& 33556\\ 
\# relations (512-split documents)&  3099  & 4330   &  3461&3610&  3063  &  4161  & 3681   & 4533   & 29938\\
\hline
Added documents due to 512-split    &  3  & 38   &  45&94& 53  & 116  & 40   & 84   & 473\\
\hline
\hline
\multicolumn{10}{|l|}{Test set} \\

\hline
\# relations (full documents)     &  814  &1728  &1208 & 1215& 1281&1597 &1299 &1933 &11075\\ 
\# relations (512-split documents)&  814  & 1559  & 1118&1043&1117  & 1294   & 1192  & 1509 &9646\\
\hline 
Impact on F1 with all GT relations&  0.0  & -3.5   & -2.26  & -5.0  & -3.8  & -5.4   &-2.6 &  -4.8 & \\
\hline 
\end{tabular}

\caption{ Impact of the 512 split on the number of considered relations (train set). Overall 12\% of the relations are lost. XFUND contains 1192 documents. The last row indicates the impact on the F1 score considering all relations in the GT and not only those retained after the 512 cut as done by \citet{Xu2021LayoutXLMMP}. Official monolingual setting with label and HQA. We simply took the correct number of initial relations to update the recall value.}
\label{512}
\end{table*}

\begin{table*}[ht]
\centering
\small
\begin{tabular}{|l|c|c|c|c|c|c|c|c|c|}
\hline
\textbf{Language}                  & \textbf{EN} &\textbf{ZH} &\textbf{JA} & \textbf{ES} & \textbf{FR} & \textbf{IT} & \textbf{DE} & \textbf{PT} \\
\hline
\hline
\citet{Xu2021LayoutXLMMP}     &  54.83  & 70.73   &  69.63& 69.96  & 63.53& 64.15  & 65.51  & 57.18  \\
LayoutXLM-Ours (5 runs)     &  50.3(1.9)  & 72.7(1.4)   &  68.5(3.0)&69.26(1.8)  & 64.8 (1.5)  & 61.4(0.6)   & 65.1(1.8) & 57.8(1.1)  \\
\hline
\end{tabular}

\caption{Reproducibility of \cite{Xu2021LayoutXLMMP}. F1 score. Official Monolingual setting. Mean and standard deviation of 5 runs (random initialisation). }
\label{replication} 
\end{table*}

\section{Appendix F: The 512 curse}
\label{sec:appendixF}
Table~\ref{512} shows the impact of the document chunking. Based on XLM-RoBERTa, LayoutXLM accepts a sequence of maximal length 512 tokens. Some documents of the dataset, longer than 512 tokens have then to be split. Relations are destroyed if the two nodes belong to two different documents. 

\end{document}